\documentclass[]{Liebert_Author}

\title{Nocturnal Eye-Inspired Liquid-to-Gas Phase Change Soft Actuator with Laser-Induced Graphene: Enhanced Environmental Light Harvesting and Photothermal Conversion
} 

\author{Maina Sogabe$^{1,\dagger}$, Youhyun Kim $^{1,\dagger}$, Hiroki Miyazako$^{1}$ and Kenji Kawashima$^{1,*}$\\
{$^{1}$Department of Information Physics and Computing, }\\
{Graduate School of Information Science and Technology,}\\
{The University of Tokyo,}\\
{The University of Tokyo,}\\
{$\dagger$ Equally contribution }\\
{$^\ast$To whom correspondence should be addressed;}\\
{E-mail:  kenji.kawashima@ipc.i.u-tokyo.ac.jp.}
}

\begin{document} 

\maketitle


\begin{abstract}
Robotic systems' mobility is fundamentally constrained by their power sources and wiring requirements. While electrical actuation systems have achieved autonomy through battery power and wireless control, pneumatic actuators remain tethered to air supply sources. Liquid-to-gas phase change actuators utilizing low-boiling-point liquids offer a potential solution, though they typically require substantial thermal input through heating elements that maintain electrical dependencies.
External heat sources, particularly light energy, present an alternative for terrestrial applications. However, despite their optical transparency, silicone-based materials have a high volumetric heat capacity and low thermal conductivity, which limits efficient photothermal energy transfer. Previous attempts to address this issue through the incorporation of graphene or metallic powder have compromised material properties, including reduced transparency and altered elastic moduli.
Inspired by the tapetum lucidum structure found in the eyes of nocturnal animals, which enables efficient light utilization in low-light conditions, this study proposes a novel anisotropic bilayer soft actuator incorporating Laser-Induced Graphene (LIG) on the inner surface of the light-irradiated silicone layer. This creates an anisotropic structure with enhanced photothermal conversion capabilities while maintaining the advantageous properties of silicone.
Comparative analysis demonstrates that the proposed actuator exhibits significantly higher photo-induced bending efficiency than conventional silicone-based actuators. The response time improved by 54\%, decreasing from 142 seconds for pure silicone to 65 seconds, with recovery response time showing a 48\% improvement. This design maintains the silicone's transparency and flexibility while
utilizing LIG, which can be fabricated under ambient conditions,
facilitating manufacturing and diverse applications.

\end{abstract}

%%word limitation4000 word

%%%温度の書き方 60${}^\circ$C

\section{Introduction}

 Currently, most robotic actuators rely on electrical energy. For example, although pneumatic actuators use air as the driving source, they rely on electrical energy to control and supply pressed air \citep*{Pneumatic1, Pneumatic2}. Therefore, they cannot be considered completely independent of electrical energy.

For centuries, people have drawn upon a wide array of energy sources to enhance their lives. Exploring alternatives to electrical power for driving soft robots can greatly broaden the range of actuation techniques available in robotics.

Several methods have been proposed for using heating methods, such as laser irradiation \citep*{Laser, Laser1}, microwave \citep*{Microwave}, heatsink \citep*{Heatsink} and electromagnetic \citep*{Magnetic, Magnetic1}.
Kim et al. developed an electrothermal pneumatic actuator using nanofiber mats, achieving a lightweight, soft actuator that can operate without external devices by utilizing liquid evaporation \citep*{solo2}. However, this approach faces issues such as high power dependency and slow response times.

Various proposals have been made, starting with actuators like pouch motors \citep*{Altm, Niiyama, Hirai, Uramune} that utilize liquid-to-gas phase change, including pneumatic actuators driven by ambient temperature \citep*{Narumi}. 
However, these actuators operate relatively slowly, with many functioning at frequencies below 0.1Hz. This is due to significant losses in converting environmental energy into driving energy, heat transfer, and thermal conductivity issues.
Sogabe et al.\citep*{SOGABE2023114587} developed a small soft actuator that operates in environments rich in thermal energy, such as warm water. This innovation improved the actuation time for bending to around one second using only environmental energy. However, this actuator leverages the thermal properties of the surrounding water (specific heat capacity of 4186~$\rm{J/kg \cdot K}$\citep*{water}) and is thus limited to underwater use.

For broader environmental applications, systems capable of operating on land are required. Thermal energy available in natural environments includes not only water temperature or body heat\citep*{ad_ref1} but also ambient thermal gradients, which have been utilized for shape memory actuation and energy harvesting in soft robotic systems\citep*{ad_ref2}. In addition to these, other strategies harness photothermal conversion using sunlight as an energy source.

A solar-driven approach utilizes a selective absorber film and a low-boiling-point fluid to generate internal pressure through sunlight-induced evaporation, enabling untethered soft robotic actuation \citep*{Solar}. Shao et al. proposed an actuator that absorbs environmental light energy through blackbody radiation by mixing graphite into silicone, converting it into thermal energy, and driving it using the liquid-to-gas phase change (methanol)\citep*{Shao}.

However, using this approach leads to reduced actuator transparency because of the graphite content. When powders such as graphite are incorporated, they decrease Young's modulus and impose limitations on both the actuator's form and performance capabilities. For this reason, developing an efficient environmental energy absorption mechanism becomes essential - one that allows even highly transparent silicone actuators to function effectively. Our work presents bio-inspired mechanisms designed to efficiently capture environmental energy for actuator operation.

\subsection*{Bio-inspiration for efficient use of environmental energy}
Nocturnal animals rely on extremely low ambient light from the moon and stars to hunt and avoid predators. To maximize the capture of this scarce light, many nocturnal species possess a specialized reflective layer behind the retina called the tapetum lucidum (Fig.~\ref{fig1}A). By reflecting photons that have passed through the photoreceptors back into the retina, the tapetum lucidum significantly enhances photoreceptor stimulation under dim conditions.

Conventional silicone-layer-only liquid-to-gas phase-change actuators absorb only a fraction of incident light, resulting in inefficient photothermal conversion. To improve light utilization efficiency, we use laser-induced graphene (LIG) to absorb light transmitted through the silicone layer and convert it into heat, analogous to how the tapetum lucidum recovers and reuses light that would otherwise be lost.

LIG has become an important area of research in direct laser writing techniques. In this process, organic materials like polyimide are converted into porous, three-dimensional graphene-like structures through rapid thermal carbonization (pyrolysis) using $\rm{CO_{2}}$ laser irradiation \citep*{Lin2014, Chyan2018}. The resulting porous network efficiently absorbs incident photons across a broad spectrum and converts them to thermal energy without requiring external wiring or electrical connections.

Due to its high electrical conductivity and patternability, LIG has attracted considerable interest in flexible electronics and multifunctional sensing devices \citep*{you2020laser,wang2022laser}. While most research has concentrated on electro-responsive or sensing applications, these studies reveal the expanding capabilities of LIG-based platforms. Wang et al. proposed light-driven robots utilizing the photothermal Marangoni effect \citep*{wang2021laser}. Furthermore, Dallinger et al. transferred LIG onto PDMS-based hydrogel sheets and achieved rapid, humidity-driven bending under light irradiation \citep*{dall2021}. These properties make LIG a promising candidate for bioelectronics and soft actuator applications \citep*{Xu2021}.

Building on these advances, we extended LIG's application to liquid‐to‐gas phase‐change actuators. To enable efficient light‐driven actuation, we added a LIG layer to conventional silicone soft actuators \citep{SOGABE2023114587}, creating an anisotropic bilayer structure in which the LIG "tapetum" lies on the light‐incident surface beneath highly transparent silicone. This design captures any light that passes through the silicone and converts it to thermal energy, which is then transferred to the internal low‐boiling‐point fluid, causing vaporization and controlled bending of the silicone actuator.

In this study, we detail the fabrication process of the proposed actuator and evaluate its performance characteristics. Finally, to demonstrate practical applicability, we developed a multi‐finger robot in which multiple LIG‐enhanced actuators operate synchronously in response to light stimuli.

\section{Concept and fabrication process}
\subsection{Concept}
\begin{figure}
 \includegraphics[scale=0.3, angle = 270]{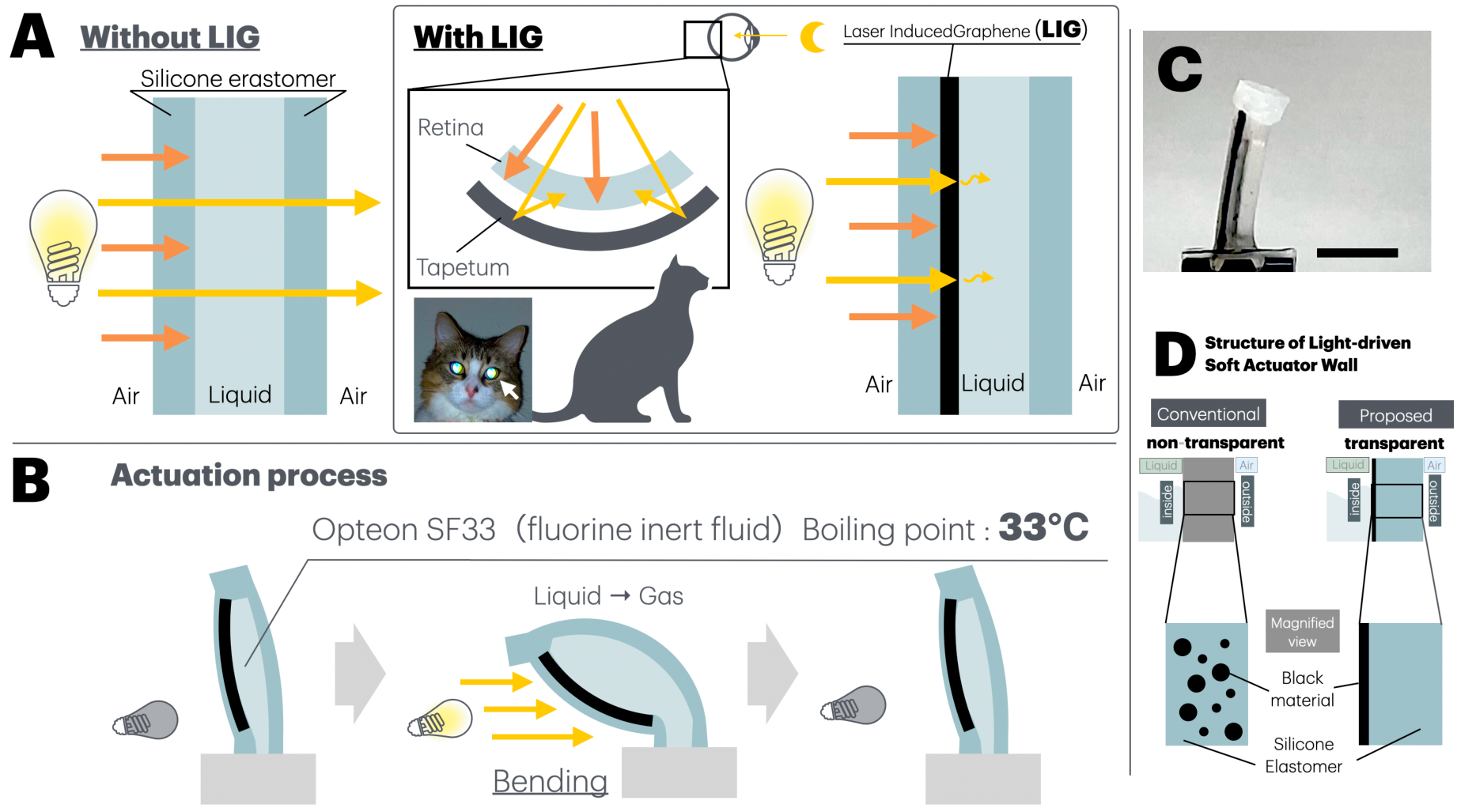}
      \caption{Concept and background of the proposed actuator.
A: Structure of tapetum lucidum in nocturnal animals' eyes and its application to the proposed actuator concept utilizing enhanced light absorption efficiency and photothermal conversion as driving force. The inset shows a photograph of the feline tapetum lucidum (indicated by the white arrow).
B: Schematic illustration of the actuator's driving mechanism and operational principles.
C: Photograph of the fabricated actuator prototype. Scale bar = 10 mm.
D: Comparative analysis between conventional soft actuators utilizing black materials (e.g., graphene) for photothermal conversion and the proposed actuator design, highlighting key differences.} 
  \label{fig1}
 \end{figure}

Figure~\ref{fig1} illustrates the conceptual design of our proposed actuator. Conventional silicone‐layer‐only liquid‐to‐gas phase‐change actuators absorb only a fraction of incident light, leading to inefficient photothermal conversion. By analogy to the tapetum lucidum's function of recovering and reusing otherwise lost light, our novel approach adopts a layered wall architecture to maximize light absorption. Specifically, we incorporate a Laser‐Induced Graphene (LIG) layer on the light‐incident surface of a highly transparent silicone elastomer (Fig.~\ref{fig1}A). This LIG "tapetum" captures any light that passes through the silicone, converting it into thermal energy via photothermal conversion to efficiently vaporize and expand the internal low-boiling-point liquid.

The actuation mechanism of the proposed device is depicted in Fig.~\ref{fig1}B. Under external light irradiation, the LIG layer heats up and transfers thermal energy to the internal working fluid (Opteon SF33, Chemours Company; $\rm{Cis\text{-}CF_{3}CH=CHCF_{3}}$), causing it to vaporize. Because the silicone walls are purposely asymmetric—thinner on the LIG side—the actuator bends directionally toward the LIG layer when the fluid expands. The operating temperature is 33${}^\circ$C (above room temperature, 25${}^\circ$C), selected to match the boiling point of the chosen fluorinated liquid.

Figure~\ref{fig1}C shows a photograph of the fabricated actuator, which measures 20 mm in length with a flattened elliptical cross-section of approximately 5 mm by 4 mm. As illustrated in Fig.~\ref{fig1}D, our bilayer design achieves efficient heat transfer without embedding additional conductive or photothermal fillers into the silicone walls. 

The LIG layer is applied to only a small portion of the actuator's interior, preserving transparency across most of the device's area. This approach leaves most of the silicone structure clear while still taking advantage of LIG's heat-generating properties when exposed to light.  This design maintains the actuator's lightweight and flexible nature while leveraging LIG's photothermal properties and facilitating easy fabrication under ambient conditions.

\subsection{Examination of LIG layer characteristics}
\subsubsection{LIG Transfer Process onto Silicone}

The LIG patterns are generated on Kapton tape (3M Polyimide Film Tape 5413, 3M Company, Minnesota, USA) adhered to a slide glass. The polyimide base of the Kapton tape enables LIG formation through laser irradiation. We utilized a laser system (xTool F1, Makeblock Co., Ltd., Guangdong, China), which is equipped with a 455~nm blue diode laser (rated power: 10~W). For LIG fabrication, the laser was operated at 40\% power with a scanning speed of 200~mm/s in unidirectional mode \citep*{miyazako}.

The LIG sheet was transferred onto a silicone (Ecoflex 00-45 Near Clear, Smooth-On, Inc., Pennsylvania, USA). For the transfer process, the LIG sheet, which remained adhered to the slide glass on one side, was positioned and surrounded by a mold. The degassed uncured silicone was then poured into the mold using a vacuum degassing chamber. The curing process was carried out at 60${}^\circ$C for 60 minutes. After curing, the mold was separated from the slide glass, and the cured silicone was peeled off the Kapton tape, successfully transferring the LIG onto the silicone surface and yielding a silicone substrate with an integrated LIG layer.

\subsection{LIG-transferred soft actuator fabrication process}
\begin{figure}
 \centering
\includegraphics[scale=0.26, angle = 270]{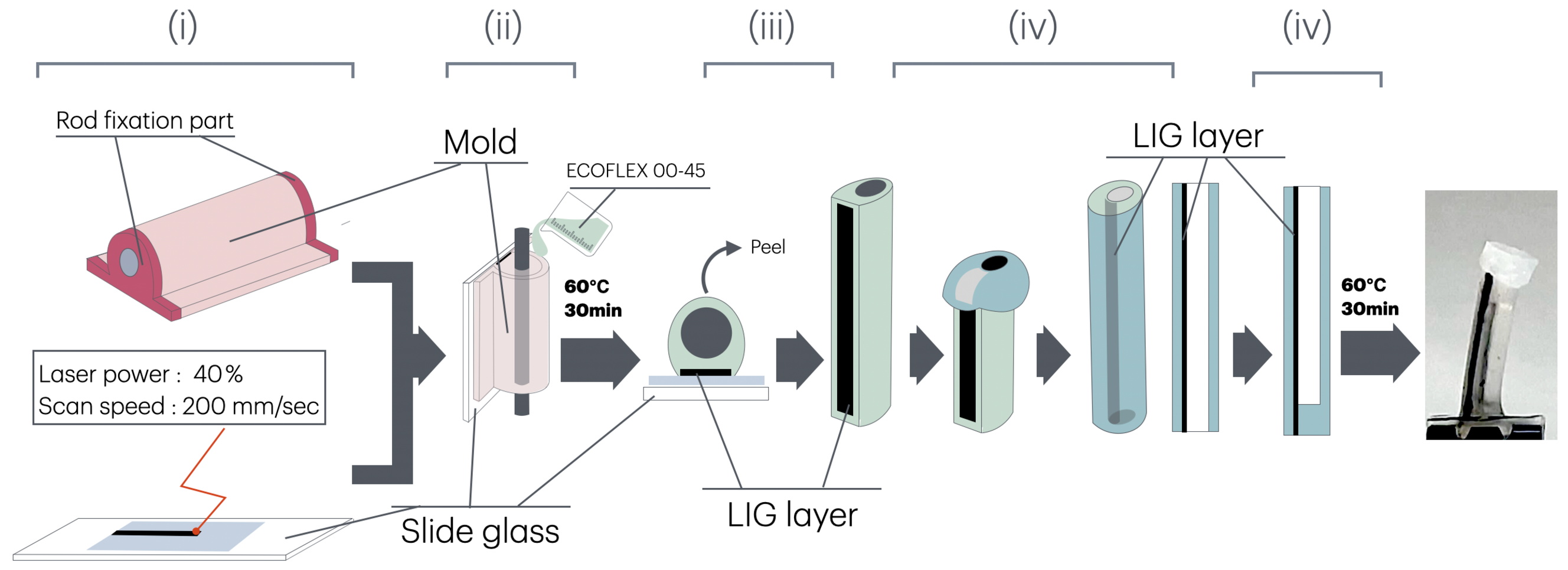}
\caption{Fabrication process of the proposed actuator. In step (i), a laser-processed plate with LIG printing on slide glass is integrated with a mold component to form the tubular structure. Subsequently, in step (ii), a metallic rod (2.5~mm in diameter) is inserted and secured at the rod fixation port, followed by the injection of degassed Ecoflex 00-45. After curing at 60${}^\circ$C for 60 minutes, step (iii) involves carefully removing the mold and peeling the silicone tube from the plate. At this stage, the LIG layer is positioned on the exterior surface; therefore, in step (iv), the tube is inverted. The fabrication process concludes with step (v), where one end of the open-ended tube is sealed with uncured silicone. The actuator becomes operational upon encapsulation with Opteon SF33 and securing the open end with a silicone-coated clip.} 
      \label{fig2}
   \end{figure}

Figure~\ref{fig2} illustrates the fabrication of a soft actuator with the LIG layer next to the liquid interface, consisting of five steps (i–v). First, a LIG pattern (2.6~mm $\times$ 40~mm) is created with the specified laser parameters. Next step, as shown in Fig.~\ref{fig2}(i), the mold and the printed LIG pattern are superimposed and fixed in place.  Next, a metallic rod (2.5~mm diameter) is placed in the mold, and silicone is poured and cured (Fig.~\ref{fig2}(ii)). In Fig.~\ref{fig2}(iii), the tubular silicone structure with integrated LIG is removed. Since the LIG is on the exterior, the tube is inverted to position LIG inside (Fig.~\ref{fig2}(iv)). In Fig.~\ref{fig2}(v), uncured silicone seals one tube end, yielding an actuator with an open terminus. Finally, 40~\textmu L of low-boiling-point liquid (Opteon SF33, Chemours, USA) is added to the cavity, and the end is sealed with a silicone-coated clip.

 Additionally, the silicone sheets (10~mm square with transferred LIG patterns) were fabricated using the same process with molds adjusted to achieve a thickness of 1.0~mm. These molds were fabricated using a Formlabs Form 3 printer (Formlabs Inc., MA, USA). Clear Resin v4 (Formlabs Inc.) was used as the mold production resin material. We evaluated temperature response to determine whether placing silicone externally and LIG internally affects photothermal efficiency. A reflector lamp (RF100 V54 WD, 100~V, 54~W; Panasonic, Japan) served as the light source. Temperature changes on the silicone sheet were measured with a thermal camera (OPTPI64 ILTO15 T090, Optris GmbH,Berlin, Germany), and illuminance was measured using an illuminance meter (LX20, Sanwa Electric Instrument Co., Ltd., Tokyo, Japan). 

Our findings revealed that positioning LIG on the interior accelerated the temperature rise at the liquid-contact interface (Supplementary Fig. 1). The initial temperature change rate during the first 5 seconds after light irradiation was 0.71~K/s when irradiated from the silicone side, compared to 0.66~K/s when irradiated from the LIG side (reference value: approximately 0.13~K/s for silicone-only sheet). This phenomenon can be attributed to silicone's high volumetric heat capacity and low thermal conductivity, which impede the efficient transfer of heat generated by LIG to the liquid-contact interface.

\subsection{Mechanical and chemical durability of LIG-transferred silicone}
\begin{figure}
 \centering
\includegraphics[scale=0.17,angle=270
]{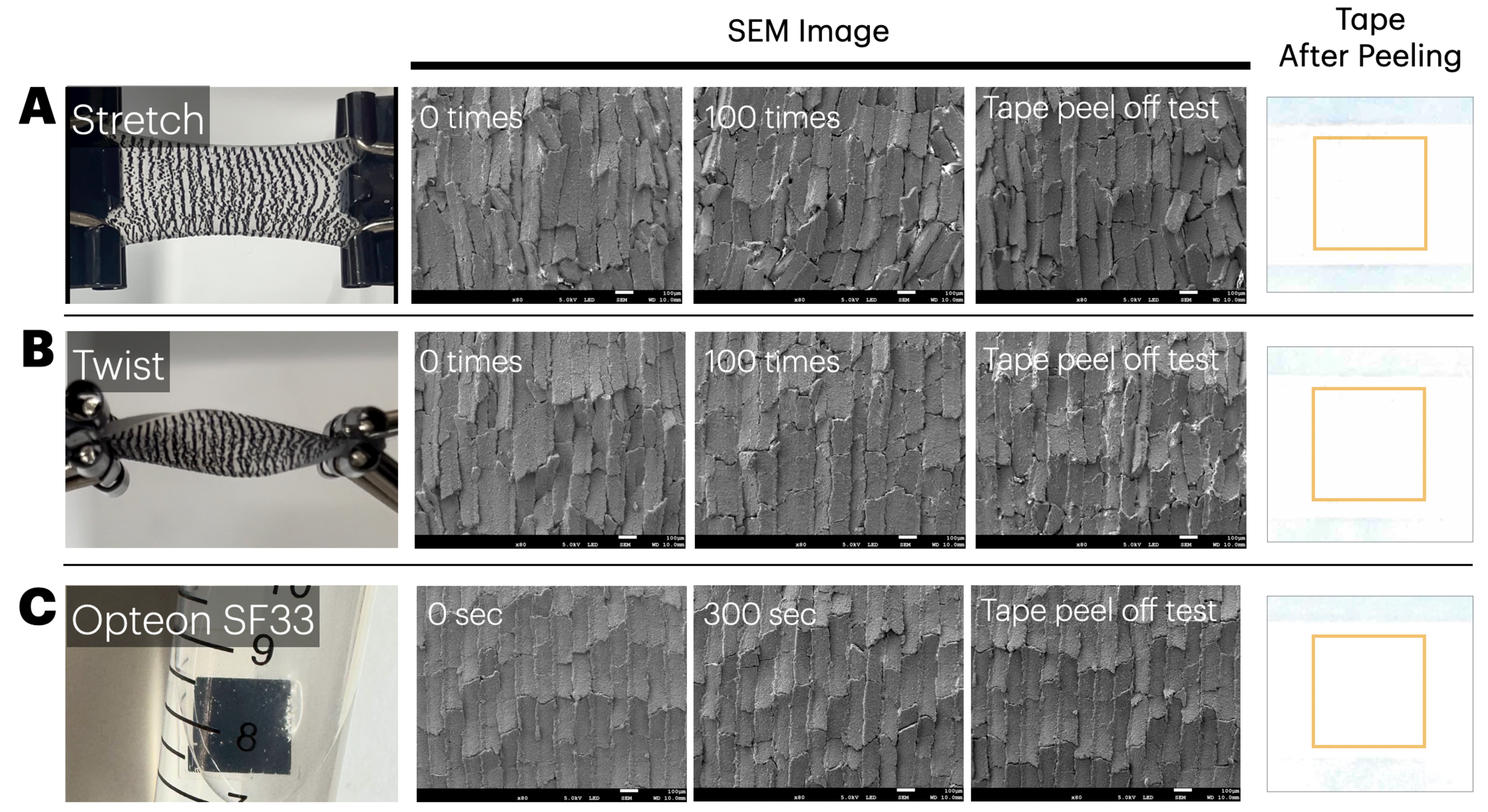}
      \caption{Durability of LIG-transferred silicone elastomer. The samples were subjected to 100 cycles of stretching to twice their original length (A), 100 cycles of 180\textdegree  twisting (B), and 300 seconds of immersion in Opteon SF33 (C). After each treatment, SEM analysis and Scotch tape tests were performed. SEM images and scanned images of the tape after peeling are shown on the right. No detachment of the LIG layer or significant morphological changes were observed in any of the conditions.
} 
      \label{fig3}
   \end{figure}

To assess the mechanical integrity and interfacial bonding of LIG transferred onto silicone elastomers, we exposed samples to operational stresses. The experimental protocol included uniaxial stretching, torsional deformation, and chemical exposure tests using Opteon SF33. Test specimens consisted of silicone elastomers with  10~mm $\times$ 10~mm LIG patterns. Samples underwent biaxial elongation to 200\% strain through 100 loading cycles (~\ref{fig3}A). Twisting testing involved $\pm$ 180\textdegree rotations repeated over 100 cycles (~\ref{fig3}B). Chemical compatibility was examined through 5-minute immersion in Opteon SF33 (~\ref{fig3}C). Adhesion was quantified using the tape peel-off test\citep*{scotch1, scotch2}. After conditioning, 3M Scotch Tape was applied, equilibrated, and then removed from the substrate surface.
Pre- and post-treatment samples were analyzed by scanning electron microscopy (SEM) to identify delamination. Removed tape sections were examined LIG debris. 

Figure~\ref{fig3} presents SEM images of LIG surfaces before and after treatment protocols. Imaging was performed using a field-emission SEM (JSM-7800F-PRIME, JEOL Ltd., Tokyo) at 5~kV accelerating voltage. Specimens received osmium coating (Tennant 20, Meiwafosis Co., Ltd., Tokyo) prior to examination. SEM analysis revealed negligible morphological alterations following stretching, torsional, or immersion treatments. The tape peel-off test produced no observable delamination or LIG structural modification, with no detectable transfer residue on tape surfaces. These findings demonstrate robust LIG–elastomer interfacial bonding sufficient for anticipated actuator operating conditions.

\section{Heat transfer analysis}

When an object is heated by radiation from a heat source at temperature $\theta_h$, the radiative heat transfer $Q_{hs}$ to the silicone wall can be expressed as
\begin{equation}
Q_{hs} = \frac{\sigma (\theta_h^4 - \theta_s^4)\,A}{\dfrac{1}{\epsilon_h} + \dfrac{1}{\epsilon_s} - 1}.
\end{equation}
Here, 
$\sigma = 5.67037442 \times 10^{-8} ~\mathrm{W\cdot m^{-2}\cdot K^{-4}}$ is the Stefan-Boltzmann constant; $\theta_s$ is the temperature of the silicone wall; $A$ is the heat transfer area; and $\epsilon_h$ and $\epsilon_s$ are the emissivities of the heat source and the silicone wall, respectively \citep*{Thermo}. 
The thermal conductivity of the silicone, $\lambda_s$, and its thickness, $x$, are given in Table~\ref{table1} \citep*{Silicone}.

The actuator's response time is sufficiently long.
Therefore, the thermal Fourier number becomes large ($F_o >>1$), and the temperature distribution within the silicone wall is negligible. The energy balance equation for the silicone wall can then be written as
\begin{equation}
C_s W_s \frac{d\theta_s}{dt} = Q_{hs} - Q_{se} \;=\; Q_{hs} \;-\; 2\,h_{se}\,A\,(\theta_s - \theta_e),
\end{equation}
where $C_s$ is the specific heat capacity of the silicone, $W_s$ is its mass ($W_s = \rho_s \, A \, x$), $h_{se}$ is the heat transfer coefficient between the silicone wall and the environment, and $\theta_e$ is the environmental temperature (25${}^\circ$C). 

Using a difference method, the following equation is derived from Eqs.~(1) and (2):
\begin{align}
\begin{split}
\theta_s(t + \Delta t) 
&= \theta_s(t) 
+ \frac{\sigma\,[\,\theta_h^4 - \theta_s(t)^4\,]\;A}{C_s \,\rho_s\,A\,x \,\Bigl(\dfrac{1}{\epsilon_h} + \dfrac{1}{\epsilon_s} - 1\Bigr)}
- \frac{2\,h_{se}\,\bigl[\theta_s(t) - \theta_e\bigr]}{C_s\,\rho_s\,x}.
\end{split}
\end{align}

When the LIG layer is deposited on the side of the silicone wall opposite the heat source, the heat transfer from the heat source to the LIG layer is
\begin{equation}
Q_{hL} 
= \frac{\sigma\,(\theta_h^4 - \theta_L^4)\,A}{\dfrac{1}{\epsilon_h} + \dfrac{1}{\epsilon_L} - 1},
\end{equation}
where $\theta_L$ is the temperature of the LIG layer and $\epsilon_L$ is its emissivity. 
The energy balance equation for the LIG layer is
\begin{equation}
C_L W_L \frac{d\theta_L}{dt} 
= Q_{hL} \;-\; h_{Le}\,A\,(\theta_L - \theta_e) \;-\; Q_{Ls},
\end{equation}
with
\begin{equation}
Q_{Ls} 
\simeq \lambda_s \,A\, \frac{\theta_L - \theta_s}{x}.
\end{equation}
Here, $C_L$ is the specific heat capacity of the LIG layer, $W_L = \rho_L\, A\,x_L$ is its mass, $h_{Le}$ is the heat transfer coefficient between the LIG layer and the environment, and $\lambda_s$ is the thermal conductivity of the silicone. 
Under these conditions, the energy equation of the silicone wall is written as
\begin{equation}
C_s W_s \frac{d\theta_s}{dt} 
= Q_{hs} \;-\; \,h_{se}\,A\,(\theta_s - \theta_e) \;+\; Q_{Ls}.
\end{equation}
The temperature of the LIG layer, $\theta_L(t)$, can be computed by a difference method based on Eqs.~(4)--(7).

The parameters used in the simulation are summarized in Table~\ref{table1}\citep*{Parameter, Parameter2}.

\begin{table}[]
\begin{threeparttable}
\caption{Parameters used in the simulation\
\label{table1}}
\setlength{\tabcolsep}{45pt}%
\begin{tabular}{@{}ll}
\toprule
 Parameter  & Value  \\
\midrule
% $\theta_h$ & $493~\rm{K}$\\

$\theta_s(0), \theta_L(0), \theta_e$ & $298~\rm{K}$\\

% $\epsilon_h, \epsilon_s$ & 0.3\\

% $\epsilon_L$ & 0.8\\

$h_{se}$ & $6~\rm{W/m^2\cdot K}$\\

$h_{Le}$ & $18~\rm{W/m^2\cdot K}$\\

$C_{s}$ & $1300~\rm{J/kg\cdot K}$\\

$C_{L}$ & $700~\rm{J/kg \cdot K}$\\

$\rho_{s}$ & $1050~\rm{kg/m^3}$\citep*{ecoflex_sds}\\

$\rho_{L}$ & $ 400~\rm{kg/m^3}$\\

$\lambda_s$ & $0.2~\rm{W/m\cdot K}$\\

$x$ & $1.0\times 10^{-3}~\rm{m}$\\

$x_{L}$ &$ 1.0 \times 10^{-4}~\rm{m}$\\
$A$ &$ 1.0\times 10^{-4}~\rm{m^2}$\\
$\alpha_s$ & $0.17$\\
$\alpha_L$ & $0.83$\\
$Q_h$ & $75~\rm{mW}$\\
\bottomrule
\end{tabular}

\end{threeparttable}
\end{table}

In this simulation, for model simplification, several factors, such as Opteon SF33 vaporization and silicone elastomer elongation, were not taken into account. The numerical values used for verification were obtained from the referenced literature. Additionally, while the radiative heat transfer equations are presented above, in our simulation, we simplified the heat transfer calculation by considering the heat source as a light source with constant energy $Q_h$. The energy absorbed by each material was then calculated using material-specific absorption coefficients($\alpha_s$ and $\alpha_L$), which are determined experimentally, where $\alpha_s Q_h$ represents the energy absorbed by the silicone wall and $\alpha_L Q_h$ represents the energy absorbed by the LIG layer. 

\begin{figure}
\centering
\includegraphics[scale=0.25,angle=270]{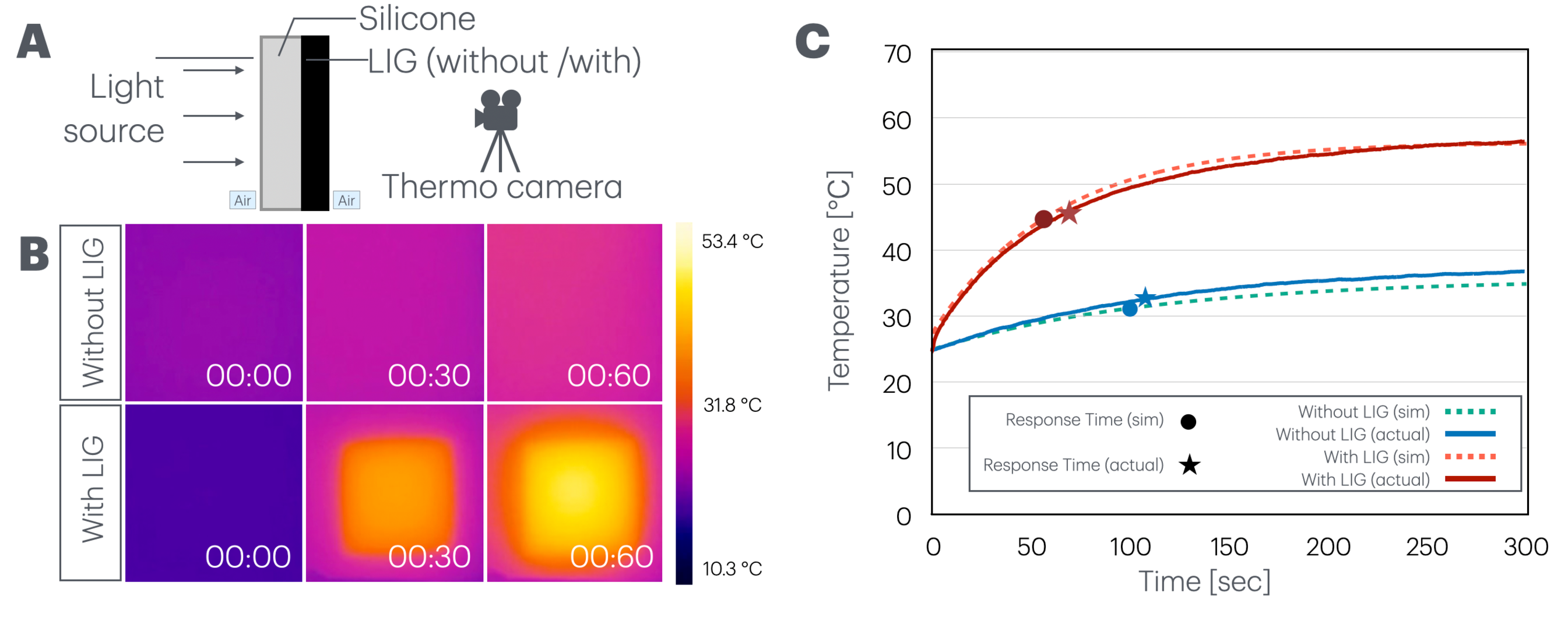}
\caption{Thermodynamic analysis of temperature changes at liquid-solid interface through thermal imaging and simulation
A: Experimental setup for thermal imaging analysis. The measurement was conducted with the distance between the light source and the silicone sheet set to 100~mm. B: Real-time thermal imaging of temperature elevation at the liquid-solid interface comparing silicone sheets with- and without-LIG. The LIG-transferred silicone sheet demonstrated more rapid temperature elevation and higher maximum temperature (56.5${}^\circ$C with-LIG vs. 36.8${}^\circ$C without-LIG). C: Comparison between experimental thermal imaging data (solid lines) and thermodynamic simulation results (dashed lines). Time constant (63\% response time) analysis revealed simulated 63\% response time values of 100.0 sec and 54.9 sec for without- and with-LIG groups, respectively, while experimental measurements yielded 63\% response time values of 103.9 sec and 62.6 sec, respectively. The experimental measurements were performed at an illuminance of 15.2 klx. Mean values are shown in the graphs (n = 3, maximum standard deviations of 0.32 and 0.57 for Without-LIG and With-LIG groups, respectively).} 
      \label{fig4}
   \end{figure}

For comparison with simulation results, we measured the temperature at the liquid contact surface using a thermal camera with silicone sheets of the same thickness as in the simulation (The experimental setup for thermal data acquisition is shown in Fig.~\ref{fig4}A, while the corresponding thermal imaging results are presented in Fig.~\ref{fig4}B). The results are shown in Fig.~\ref{fig4}C. When overlaying the temporal temperature variations with the simulation graphs, despite some discrepancies, the 63\% response time values showed close agreement: for the case without LIG, the simulation yielded 100.0~seconds while the experimental measurement was 103.9~seconds; for the case with LIG, the simulation predicted 54.9~seconds compared to the experimental value of 62.6~seconds.
These results demonstrate that LIG effectively functions as a photothermal conversion and facilitates heat transfer to the internal liquid.

The LIG–silicone actuator cooled more rapidly because it reached a higher peak temperature (56.4${}^\circ$C vs. 36.7 ${}^\circ$C), creating a larger $\delta$ T relative to the ambient (~25${}^\circ$C). According to Newton's law of cooling, this greater temperature difference increases the rate of heat loss. Furthermore, the porous LIG structure increases both surface area and thermal conductivity, which also accelerates dissipation.

\section{Evaluation of a soft actuator with transferred LIG under general light source}

\subsection{Experimental analysis of actuation driven by light On-Off cycling}

\begin{figure}
 \centering
\includegraphics[scale=0.36]{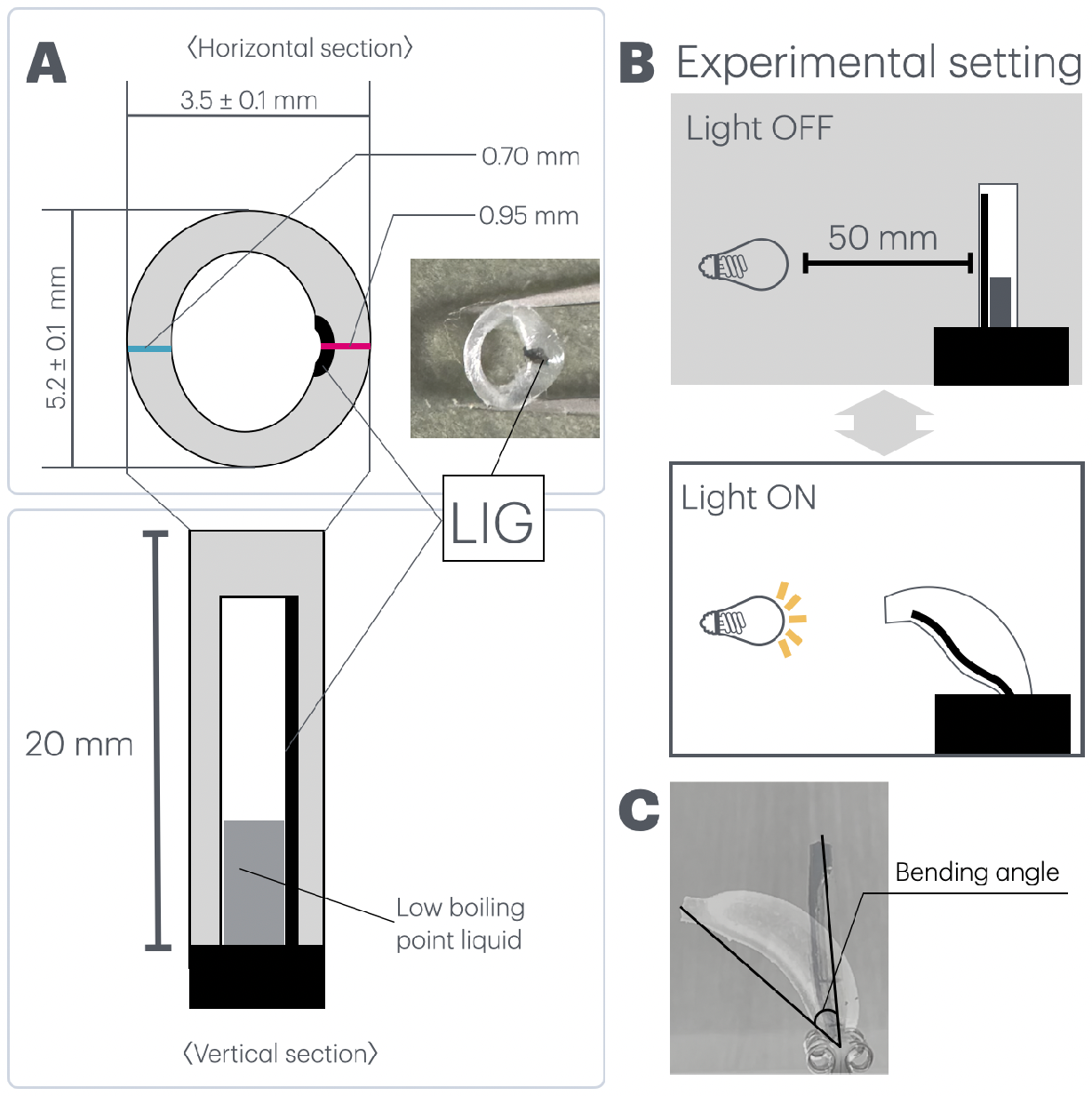}
      \caption{Overview of the proposed actuator and experimental setup.
 A: Schematic illustration and dimensions of the proposed actuator. The cross-sectional view reveals a distorted elliptical shape, with slight deformation observed at the adhesion interface with the LIG.
B: Schematic diagram of the light irradiation experimental setup. The fundamental experiments were conducted with the actuator positioned 50 mm from the light source. A 54~W reflector bulb operating at 100~V served as the light source (31.3~klx). C: Measurement method of bending angle: The initial state of the actuator before bending was defined as 0 degrees, and the angle by which the tip of the actuator tilted from this position was recorded as the bending angle.} 
      \label{fig5}
   \end{figure}

The experimental setup is illustrated in Fig.~\ref{fig5}. During experiments, the fabricated actuator was positioned 50~mm from the light source. The light bulb had a luminous intensity of 300~cd, a total luminous flux of 630~lm, and a beam angle of 60\textdegree. An iPhone 15 Pro Max (Apple Inc., Cupertino, CA, USA) captured actuator deformation, and ImageJ analyzed bending angles and deformation measurements\citep*{ImageJ}.

Figure~\ref{fig6}A, C and Supplementary Movie S1 show bending behavior during light irradiation, subsequent recovery upon light off, and corresponding analysis results. When exposed to light, the without-LIG actuator exhibited a 63\% response time of 141.7~seconds and required approximately 220~seconds to reach maximum bending. In contrast, the proposed actuator with LIG demonstrated a mean 63\% response time of 65.0~seconds and reached maximum bending in about 90~seconds, a 54.1\% improvement.

During recovery after light off (Fig.~\ref{fig6}, D and Supplementary Movie S1), the without-LIG actuator showed a 63\% response time of 42.7~seconds and needed 120~seconds to return to its initial state. The proposed actuator with LIG demonstrated a 63\% response time of 22.1~seconds and took about 60~seconds to return, a 48.2\% improvement in recovery performance.

To assess repeatability under cyclic light conditions, we performed multiple actuation tests with a single fluid charge (Fig.~\ref{fig7}). We encapsulated 40~\textmu L of Opteon SF33 and exposed the actuator to repeated light irradiation. The bending angles during the first three cycles were 51.7\textdegree, 51.6\textdegree, and 50.7\textdegree, demonstrating good consistency. In the fourth cycle, however, the bending angle declined to 90.4\% of the initial peak value and, by the sixth cycle, decreased to 50.5\%. We attribute this degradation to gas leakage through the silicone elastomer’s gas permeability. When we immersed the actuator in water at 65${}^\circ$C,, gas bubbles emerged from expanded and thinned areas of the silicone wall (Supplementary Figure 2). To address this limitation, we designed the actuator with a refillable configuration and incorporated a detachable clip-based sealing mechanism for repeated operation.

\subsection{Analysis of Actuator Performance Under Different Liquid Volume and Light Distance Conditions}

Figure~\ref{fig8} demonstrates the variations in actuator performance based on changes in the encapsulated liquid volume and the distance between the light source and actuator. With the light source fixed at 50~mm, we examined the effect of varying volumes (20~\textmu L, 30~\textmu L, and 40~\textmu L) of encapsulated Opteon SF33. The results showed volume-dependent increases in three parameters: the angular change ratio (Fig.~\ref{fig8}A), the plateau bending angle (Fig.~\ref{fig8}B), and the time required to reach the plateau state (Fig.~\ref{fig8}C). The linear regression analysis yielded an $R^2$ value of approximately 0.98, indicating that the relationship can be effectively approximated as linear within the 20--40~\textmu L range.

We examined how environmental conditions affect the actuator by testing it at different distances from the light source 50~mm, 75~mm, and 100~mm. The results are shown in Fig.~\ref{fig8}D--F using 30\textmu L of Opteon SF33. As we moved the actuator further from the light source, both the angular change ratio and the plateau bending angle became smaller. The actuator positioned at 100~mm from the light source required a significantly longer time to reach its final bent configuration, approximately 2.5 times greater than that observed at 50~mm distance.

\section{Mechanical characteristics of the actuator}
We evaluated the load-bearing performance of the actuator by attaching polyurethane sponge weights to its distal tip and measuring the resulting reduction in bending angle(Fig.~\ref{fig9}A). A nylon thread was connected to the tip of a pre-bent actuator, and sponge-based weights of 0.125, 0.25, 0.5, 0.75, 1.0, and 1.5~g were sequentially suspended via a fine pin. 
The results (Fig.~\ref{fig9}B) indicate that the actuator maintained approximately 90\% of its original bending angle under a 0.125~g load ($\approx$ 1.23~mN). As the load increased, the actuator's ability to retain its bent configuration declined, reaching around 50\% of the original angle under a 1.5~g load ($\approx$ 14.7~mN). These results demonstrate a near-linear relationship between the applied load and bending angle reduction, providing a quantitative evaluation of the actuator's mechanical output capacity.

\section{Multi-Finger Robot Using Non-Contact \& Light-Driven LIG-Transferred Soft Actuators}
Investigating the practical applications of this actuator, we conducted experimental validation of the multi-finger actuator's performance in controlled laboratory conditions and assessed its functionality under natural solar illumination.
\subsection{Three-finger actuator performance in laboratory conditions}
To evaluate the feasibility of the proposed actuator system, we investigated synchronized operation of multiple actuators under controlled light exposure conditions and examined the effects of photoactivation and deactivation on actuator performance. We assessed the coordinated bending motion of three actuators in response to a light stimulus, aiming to achieve simultaneous coverage of a designated target object. The experimental results are documented in Supplementary Movie S2 and Supplementary Figure S3. The experimental setup comprised a controlled laboratory environment. Illumination was provided by reflector lamps throughout experimental trials. 

The experimental findings demonstrated that upon exposure to incident light at an intensity of approximately 15 klx, all three actuators exhibited synchronized bending behavior with high temporal coordination. Upon cessation of light exposure, the actuators consistently returned to their initial linear configuration, confirming the reversible nature of the photoresponsive actuation mechanism.

\subsection{Solar-responsive shading device demonstration}
In natural environments, intense solar radiation can cause harm to flora and fauna. To address this challenge, we developed a prototype shading device (Fig.~\ref{fig10}) that responds directly to solar irradiance by providing shade coverage over designated target organisms. The device architecture incorporates three actuators integrated with nonwoven fabric. We deployed this system outdoors to evaluate solar-driven performance.
Testing was conducted at 14:00 JST under clear atmospheric conditions. On the testing date, the ambient illuminance measured 102.5~klx. Using clover as the target organism, we positioned the device under direct solar exposure. As demonstrated in Fig.~\ref{fig10} and Supplementary Movie S3, the actuators responded to sunlight by bending and successfully providing shade coverage over the clover. Following deployment, illuminance within the shaded region decreased to 37.6~klx, achieving a 63.3\% reduction. The actuation process required approximately 140 seconds to complete.

\section{Limitations \& future work}
\subsection{The operating temperature range control}
We employed Opteon SF33 with a boiling point of 33${}^\circ$C for the actuator's internal fluid. We hypothesize that the operating temperature range can be controlled by varying the boiling point of the encapsulated liquid. For instance, Opteon SF79 and Opteon SF70 have boiling points of 48${}^\circ$C and 71${}^\circ$C, respectively. Thermal imaging data (Fig.~\ref{fig4}B) demonstrates that temperatures approaching 55${}^\circ$C can be achieved when LIG is transferred, suggesting that for terrestrial applications, this system could be implemented with liquids having boiling points in the range of 30-55${}^\circ$C. However, conventional light sources may prove insufficient for operations requiring higher temperature ranges. In such cases, focusing lenses, high-powered lighting systems, or higher-output lasers are required.

\subsection{Gas leakage}
As shown in Supplementary Fig. 2, gas leakage phenomena were confirmed to occur in the proposed actuator, similar to previously reported issues\citep*{SOGABE2023114587}. The current actuator design features a reusable mechanism with one side sealed by a clip, allowing for multiple liquid refilling cycles. Future work requires optimal design of the refilling port to enhance sealing performance and improve airtightness.
However, gas leakage can limit potential applications. Potential approaches include incorporating materials with lower gas permeability and developing alternative sealing mechanisms to address this limitation.

\subsection{Utilization of ambient environmental light}
The optical characterization and performance evaluation were conducted under controlled laboratory conditions and outdoor environments. In the laboratory setup, a reflector lamp (54~W, beam angle: 60\textdegree) was employed as the primary light source, exhibiting a center luminous flux of 630~lm and a beam luminous flux of 160~lm. Illuminance measurements yielded approximately 15~klx at a working distance of 100~mm and 30~klx at 50~mm, respectively.

Comparative outdoor experiments under direct solar irradiation demonstrated higher illuminance levels of 100–110 klx, indicating laboratory conditions represented about 30\% of natural outdoor intensity. The actuator demonstrated slower bending speed outdoors despite increased light intensity, requiring approximately 140 seconds to reach maximum bending angle (Fig.~\ref{fig10}). This slower performance stems from convective heat transfer caused by wind, which cools the actuator and prevents the necessary temperature rise. Outdoor deployment faces challenges from seasonal changes, location factors, and weather conditions, limiting fast response times. Optical concentration systems, such as focusing lenses, could boost light energy density on the LIG surface. Smart use of wind-induced cooling might enable quicker bending–unbending cycles, broadening outdoor applicability. 

Additionally, localized light energy irradiation using laser sources to improve actuator control precision at the microscale represents a promising avenue for future research.

\section{Conclusions}
We addressed the limitation of conventional silicone-based soft actuators, which exhibited low driving efficiency due to ineffective light utilization. We incorporated a photothermal conversion layer using LIG on the light-incident surface. This modification enabled the actuator to effectively utilize light that would otherwise be poorly absorbed by silicone alone, heating the internal low-boiling-point liquid. As a result, the proposed actuator with LIG demonstrated a 54\% improvement in response time compared to conventional without-LIG actuators.

Our LIG-based approach preserves silicone's inherent characteristics while enabling fabrication under ambient conditions and operation with environmental light sources including sunlight. This method significantly simplifies the manufacturing process of photothermal actuators and opens up promising possibilities for their broader application in diverse fields.

%plainnat

\section*{Acknowledgement}
This work was supported by the Japan Society for the Promotion of Science (JSPS) KAKENHI Grant Number JP24K21213. 
This work was supported by "Advanced Research Infrastructure for Materials and Nanotechnology in Japan(ARIM)" of the Ministry of Education, Culture, Sports, Science and Technology (MEXT), Grant Number JPMXP1225UT0165.
M.S. sincerely thanks T. Kondo, who kindly provided support in the use of the scanning electron microscope, and my cats, whose eyes served as a visual reference for demonstrating the tapetum lucidum structure and whose presence inspired and supported me throughout the research process.

\section*{Author declaration}
No competing financial interests exist.

\section*{CRediT}
Maina Sogabe: Conceptualization, Data curation, Investigation, Formal analysis, Methodology, Supervision, Visualization, Writing-original draft, Writing-review \& editing

Youhyun Kim: Data curation, Methodology, Formal analysis, Resources, Writing-original draft, Writing-review \& editing

Hiroki Miyazako: Methodology, Writing-review \& editing

Kenji Kawashima: Funding acquisition, Investigation, Formal analysis, Project administration, Supervision, Writing-original draft, Writing-review \& editing

\section{Supplementary Material}

\begin{figure}
 \centering
\includegraphics[scale=0.24,angle=270]{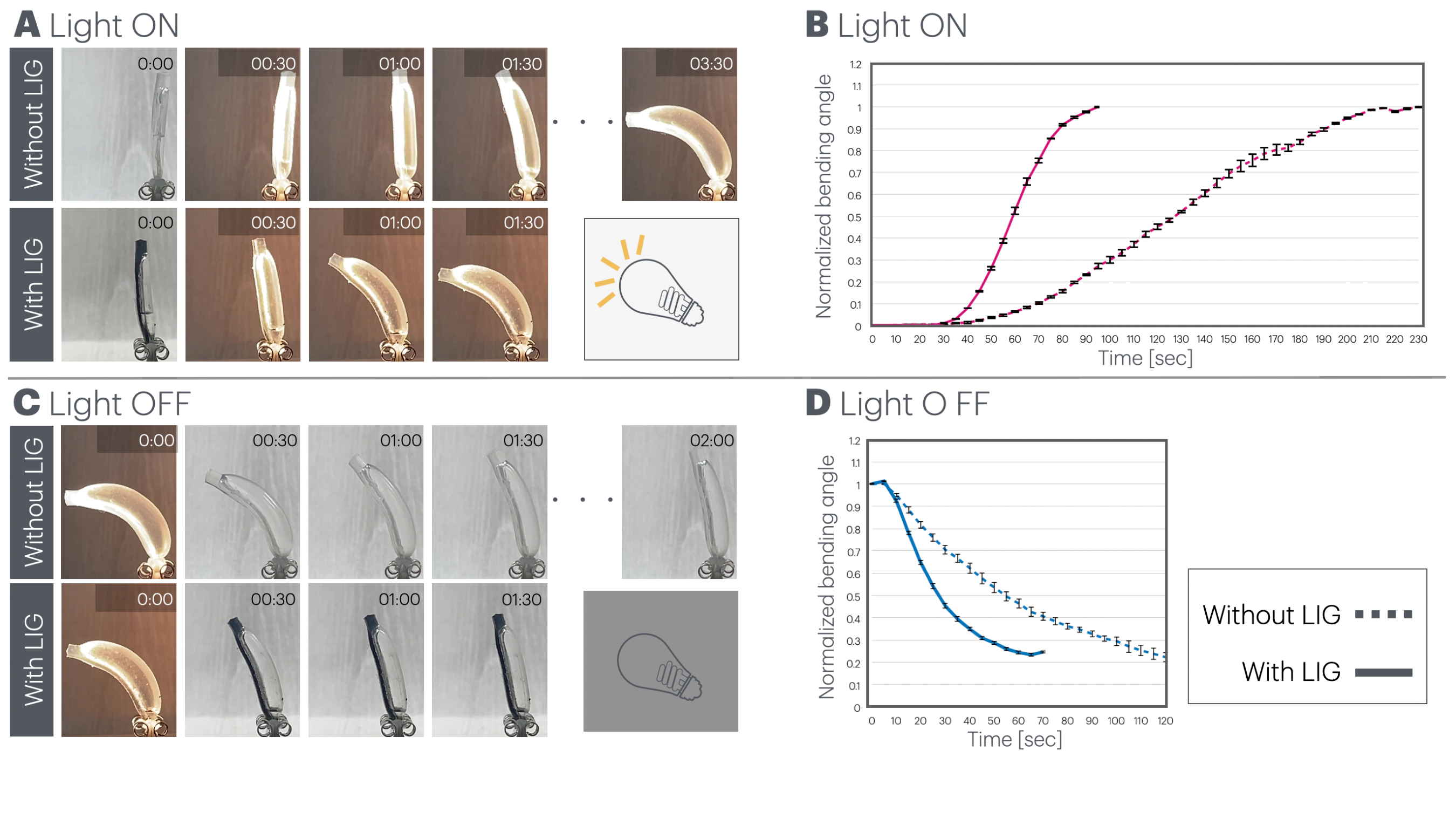}
      \caption{Temporal analysis of bending actuator dynamics. A: Time-lapse photographs showing bending deformation during light irradiation with and without LIG.
B: Graph of the normalized bending angle during light exposure (n = 5). Angles were normalized by setting the initial angle to 0 and the maximum plateau angle to 1, enabling consistent comparison across samples.
C: Time-lapse photographs illustrating the recovery process after deactivating the light source.
D: Corresponding graph of the normalized bending angle during recovery (n = 5). The plateau angle was defined as the value beyond which the angular change remained less than 1 degree over a 20-second period.
  All error bars represent the standard error of the mean (SEM).} 
      \label{fig6}
   \end{figure}

\begin{figure}
 \centering
\includegraphics[scale=0.23,angle=270]{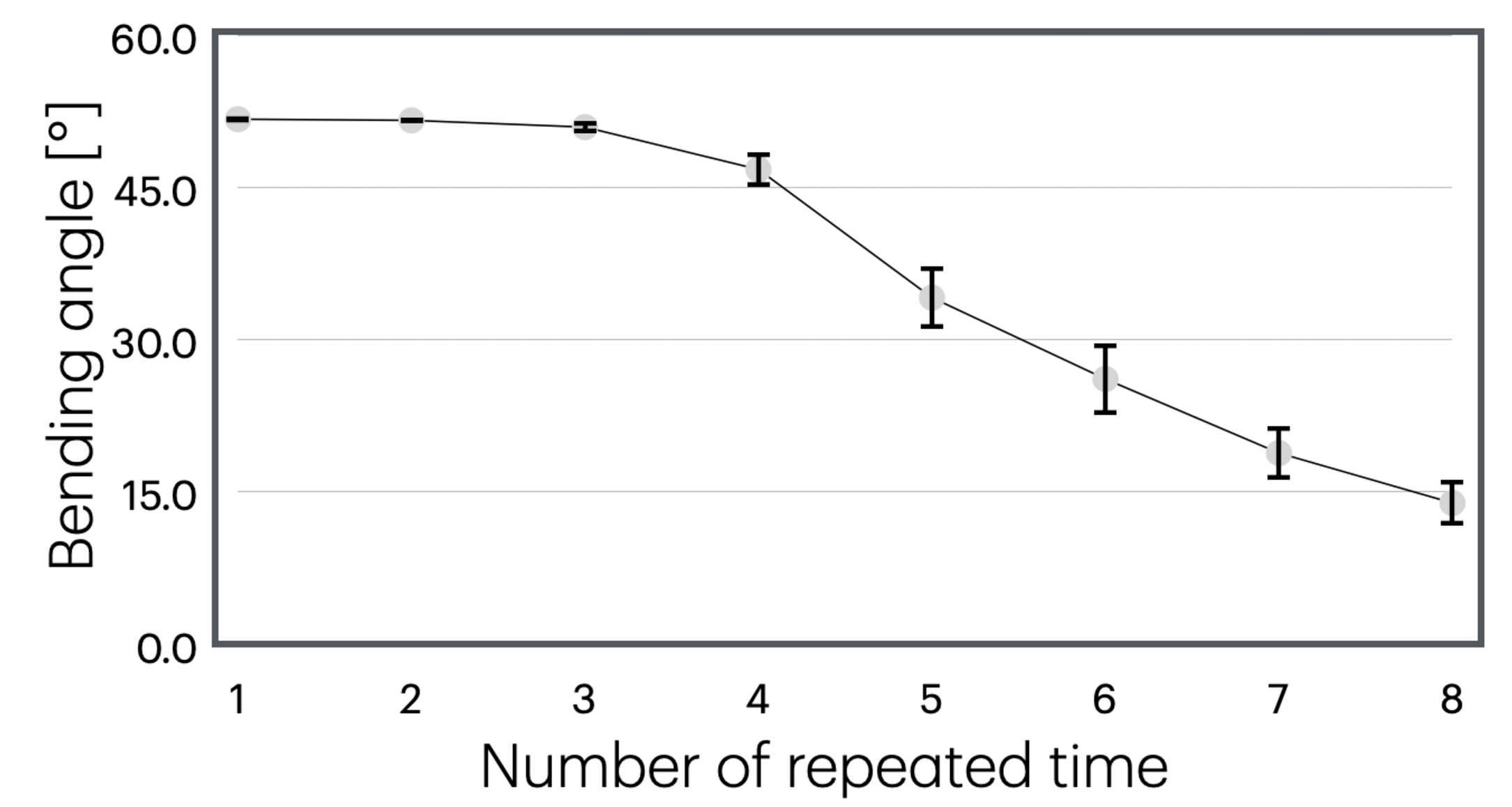}
      \caption{Change in bending angle during repeated actuation with a single liquid injection.
Variation in final bending angle over multiple light irradiation cycles using an actuator with a maximum bending capacity of 51\textdegree. The actuators were filled with 40~\textmu L of liquid (n = 3). The bending angle was defined as the final value when no change in angle was observed for over 5 seconds. Error bars represent standard deviation.} 
      \label{fig7}
   \end{figure}

\begin{figure}
 \centering
\includegraphics[scale=0.26,angle=270]{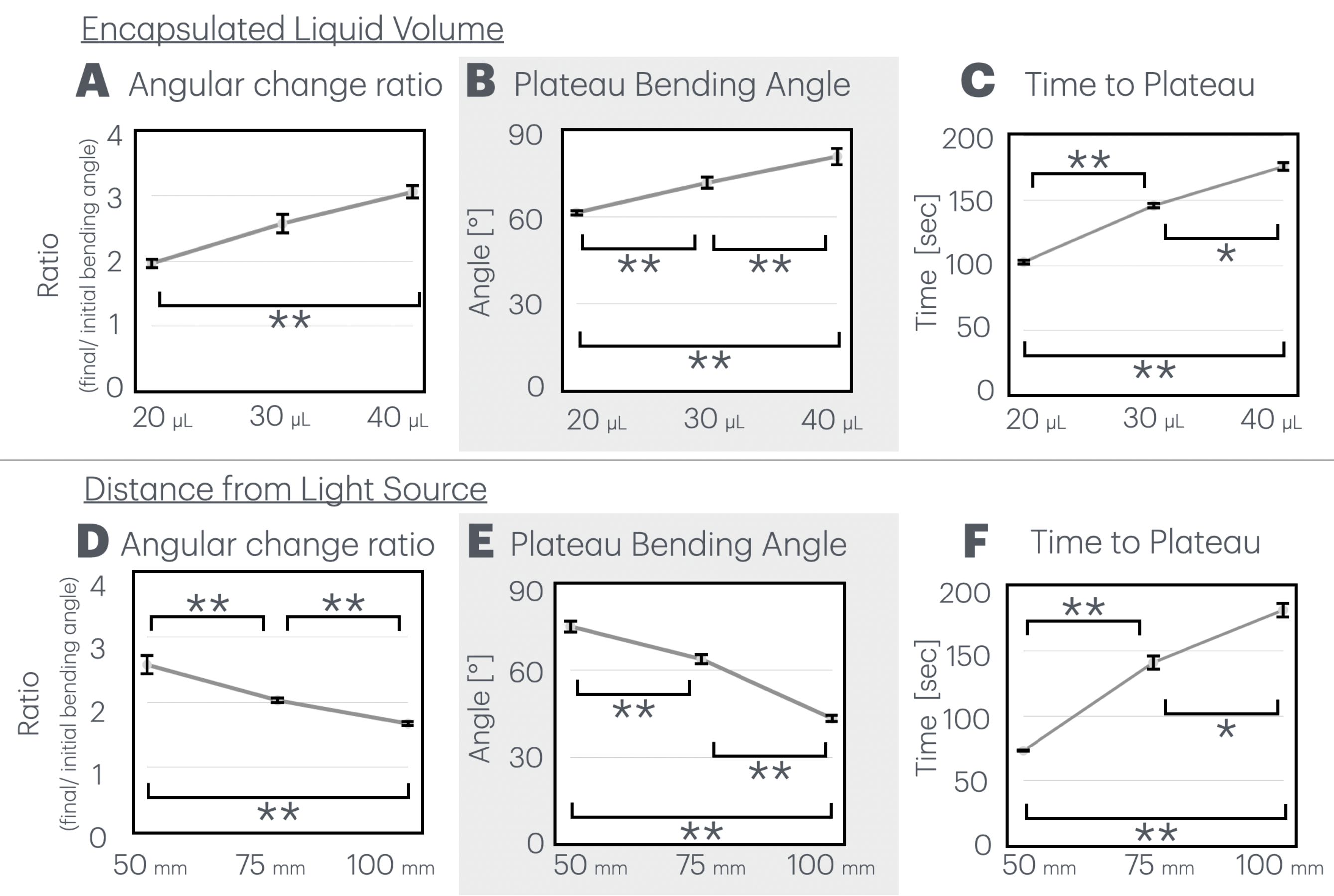}
       \caption{Analysis of LIG-transferred soft actuators under different internal and environmental conditions.  
A--C: Effects of internal liquid volume on actuator performance at a constant light source distance (50~mm). Internal liquid volumes of 20, 30, and 40~\textmu L were tested, showing (A) the angular change ratio (ratio of final to initial angle, normalized to the initial angle of 1), (B) the plateau bending angle, and (C) the time to reach the plateau state.  
D--F: Effects of light source distance on actuator performance at a constant internal liquid volume (30~\textmu L). Light source–to–actuator distances of 50, 75, and 100~mm were tested, showing (D) the angular change ratio, (E) the plateau bending angle, and (F) the time to reach the plateau state. The plateau state was defined as the condition in which the angular change did not exceed 1\textdegree over a 30-second period. Measurements were conducted using three actuators (n = 3) for both sets of experiments. Data were analyzed by one-way repeated measures ANOVA, and post-hoc comparisons between conditions were performed using the Benjamini--Hochberg method (*: $p<0.05$; **: $p<0.01$). Graphs show average values (error bars represent standard deviation).%
} 
      \label{fig8}
   \end{figure}

\begin{figure}
 \centering
\includegraphics[scale=0.27,angle=270]{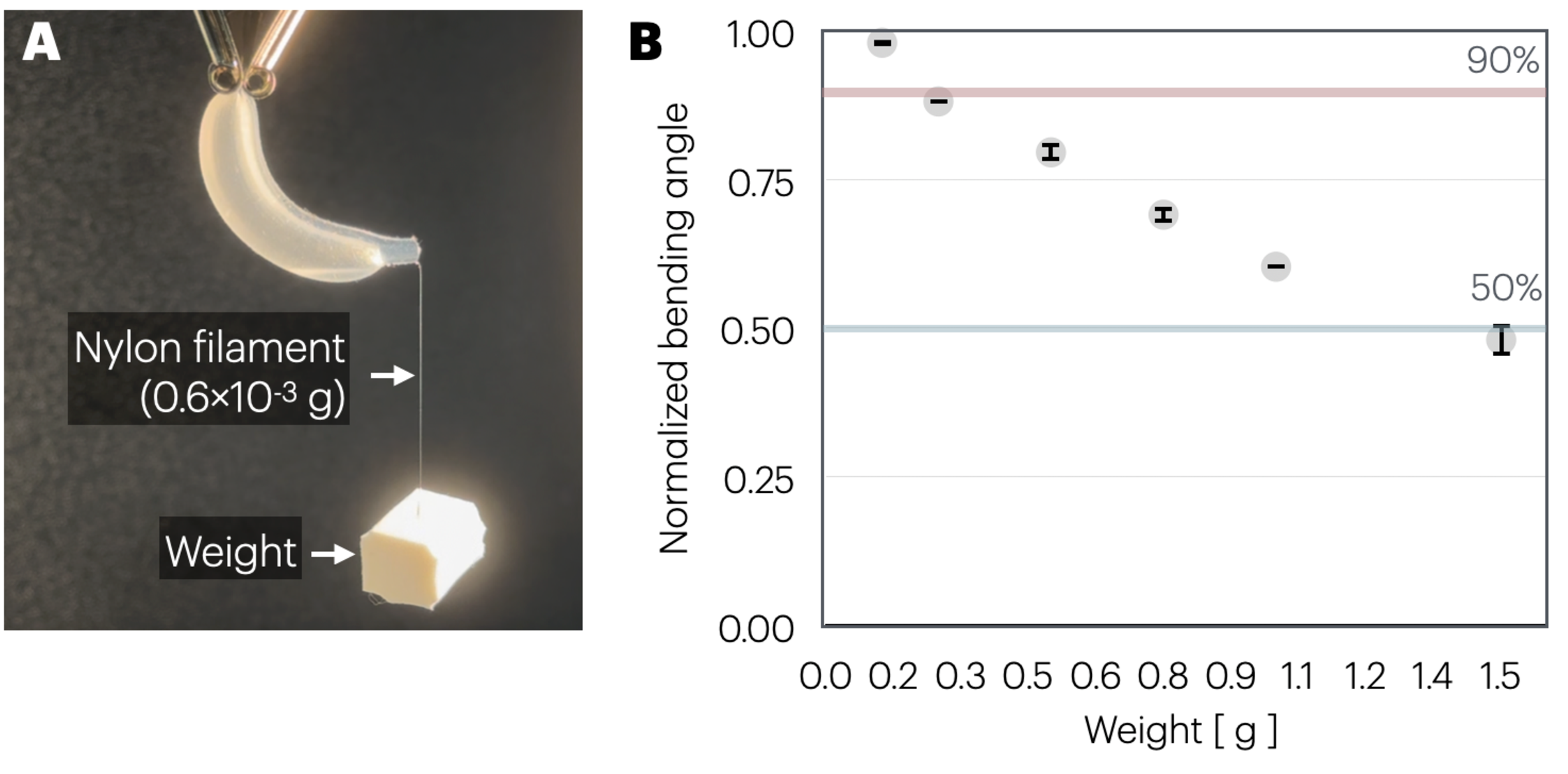}
       \caption{Load-bearing performance evaluation of the proposed actuator. A: Experimental setup showing the actuator with attached polyurethane sponge weight suspended via nylon thread. The actuator demonstrates measurable bending under load while maintaining structural integrity. B: Quantitative analysis of normalized bending angle as a function of applied load (0.125 - 1.5~g, equivalent to 1.23 - 14.7~mN). 40~\textmu L of Opteon SF33 was sealed in each measurement. Three independent experiments were performed for each condition (n = 3), and the average values are shown in the graph (error bars represent standard deviation). } 
      \label{fig9}
   \end{figure}

\begin{figure}
 \centering
\includegraphics[scale=0.26,angle=270]{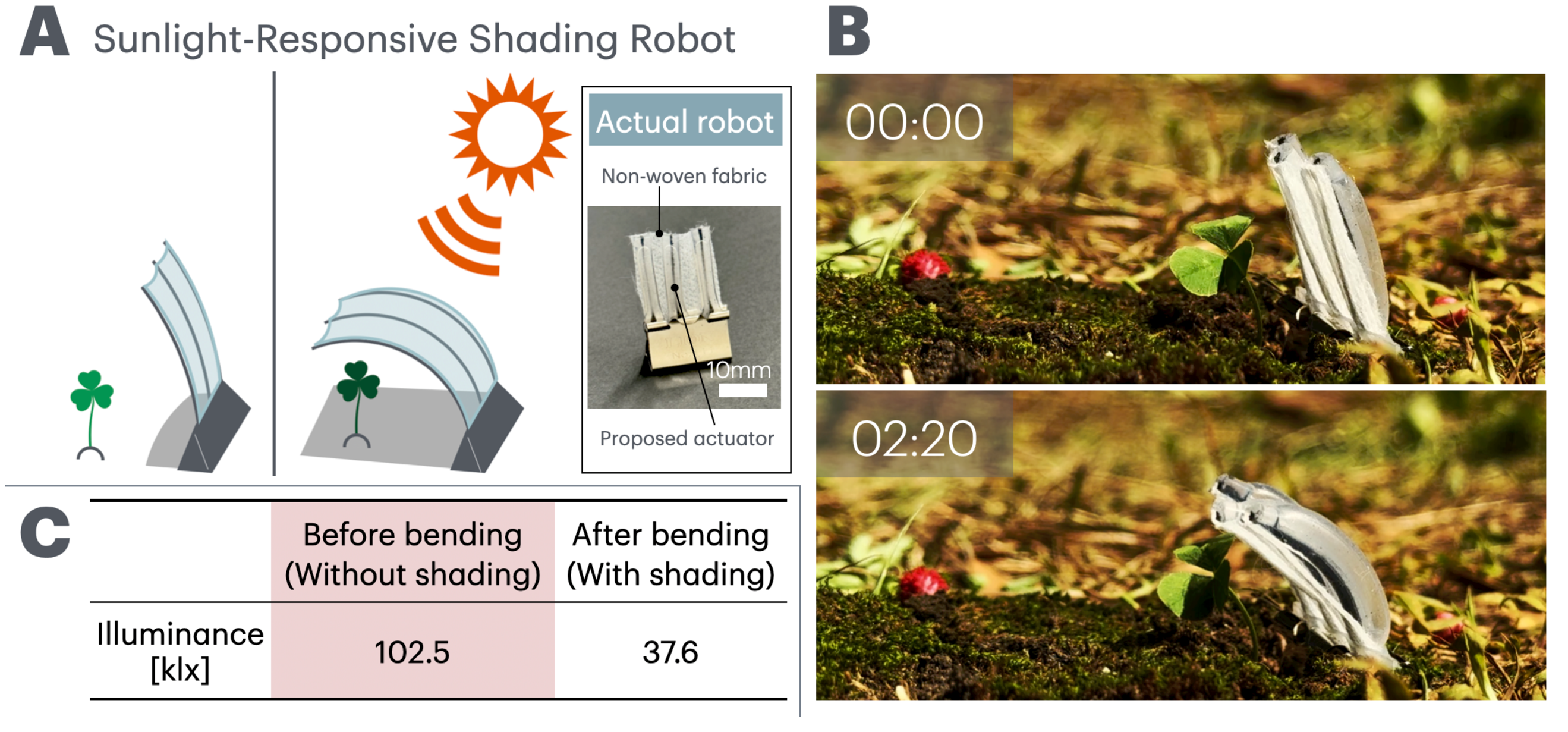}
       \caption{
       Demonstration of the proposed actuator for shading robotics. A: Overview and actual image of the robot. When exposed to sunlight, the actuator bends and blocks light around the target area. B: Operation of the actuator under natural sunlight. The actuator successfully shaded the area where the clover was present through bending motion. C: Illuminance measurements of the area with and without shading, where clover was present. The measurements were taken at 14:00 JST under clear skies, at an ambient temperature of 27${}^\circ$C. Location: 35.7139165\textdegree N, 139.7605257\textdegree E.} 
      \label{fig10}
   \end{figure}

\begin{enumerate}
\item Supplementary Figure S1:Effect of light irradiation direction
Temperature changes on the opposite surface (liquid-contact surface) when the light was irradiated from one direction onto either the silicone side (A) or LIG side (B) of silicone elastomer with transferred LIG or silicone only (C). The fastest temperature rise was observed when light was irradiated from the silicone side (case A).

\item Supplementary Figure S2: Visualization of gas leakage from the proposed actuator.
The proposed actuator was immersed in 65${}^\circ$C hot water to visualize gas leakage from within. Small bubbles can be seen leaking from the expanded actuator.

\item Supplementary Figure S3: Synchronized actuation of three-finger actuator system under light control. A: Schematic illustration showing the operating principle: actuators remain straight when light is OFF and bend to form a protective enclosure when light is ON through a "hold and protect" mechanism. B: Time-lapse demonstration of synchronized actuation in laboratory conditions showing three actuators simultaneously bending to cover a target object upon light exposure (15~klx). At t = 0.00~seconds (light-off), the actuators maintain a straight configuration. At t = 1.00~seconds (light-on), the actuators exhibit a coordinated bending motion, forming a protective enclosure. At t = 2.20~seconds (light-off), the actuators return to their initial straight state, demonstrating reversible and controllable actuation through light activation and deactivation.

\item Supplementary Movie S1: The actuator's bending under light irradiation and its bending release upon turning off the light (related to Fig.~\ref{fig6}).
\item Supplementary Movie S2: Demonstration of a multi-finger robot in a laboratory setup mimicking a heterogeneous natural environment. The ambient temperature was 20${}^\circ$C, and the illuminance was 30.2~klx. Under light irradiation, the three actuators moved synchronously, and upon light-off, their bending was released (see Supplementary Figure S3). 

\item Supplementary Movie S3: Demonstration of shading robot driven by sunlight in nature (related to Fig.~\ref{fig10}).
\end{enumerate}

%\begin{thebibliography}{99}
\bibliography{sample}
\bibliographystyle{van_ma}

\newpage
%%%-----%%%Figure

\end{document}